\documentclass[10pt,twocolumn,letterpaper]{article}

\usepackage[pagenumbers]{cvpr} % To force page numbers, e.g. for an arXiv version

\usepackage{graphicx}
\usepackage{amsmath}
\usepackage{amssymb}
\usepackage{booktabs}

\usepackage{circledsteps}
\usepackage[accsupp]{axessibility}

\usepackage[pagebackref,breaklinks,colorlinks]{hyperref}

\usepackage[capitalize]{cleveref}
\crefname{section}{Sec.}{Secs.}
\Crefname{section}{Section}{Sections}
\Crefname{table}{Table}{Tables}
\crefname{table}{Tab.}{Tabs.}

\def\cvprPaperID{11} % *** Enter the CVPR Paper ID here
\def\confName{CVPR}
\def\confYear{2022}

\begin{document}

%%%%%%%%% TITLE - PLEASE UPDATE
\title{Interactive Segmentation and Visualization for \\Tiny Objects in Multi-megapixel Images}

\author{Chengyuan Xu$^1$
	\hspace{0.15in}
	Boning Dong$^1$
	\hspace{0.15in}
	Noah Stier$^1$
	\hspace{0.15in}
	Curtis McCully$^2$ \\
	D. Andrew Howell$^{1,2}$
	\hspace{0.15in}
	Pradeep Sen$^1$
	\hspace{0.15in}	
	Tobias Höllerer$^1$\\
	$^1$University of California, Santa Barbara \hspace{0.4in}$^2$Las Cumbres Observatory \\
	{\tt\small \{cxu, boning, noahstier, psen, thollerer\}@ucsb.edu, \{cmccully, ahowell\}@lco.global}
}

\maketitle

%%%%%%%%% ABSTRACT
\begin{abstract}

We introduce an interactive image segmentation and visualization framework for identifying, inspecting, and editing tiny objects (just a few pixels wide) in large multi-megapixel high-dynamic-range (HDR) images. Detecting cosmic rays (CRs) in astronomical observations is a cumbersome workflow that requires multiple tools, so we developed an interactive toolkit that unifies model inference, HDR image visualization, segmentation mask inspection and editing into a single graphical user interface. The feature set, initially designed for astronomical data, makes this work a useful research-supporting tool for human-in-the-loop tiny-object segmentation in scientific areas like biomedicine, materials science, remote sensing, etc., as well as computer vision. Our interface features mouse-controlled, synchronized, dual-window visualization of the image and the segmentation mask, a critical feature for locating tiny objects in multi-megapixel images. The browser-based tool can be readily hosted on the web to provide multi-user access and GPU acceleration for any device. The toolkit can also be used as a high-precision annotation tool, or adapted as the frontend for an interactive machine learning framework. Our open-source dataset, CR detection model, and visualization toolkit are available at \url{https://github.com/cy-xu/cosmic-conn}.

\end{abstract}

%%%%%%%%% BODY TEXT

% Submission of papers describing demonstrations
% A paper submitted to accompany a demonstration should outline the design of the system and provide sufficient details to allow the evaluation of its validity, quality, and relevance to CVPR. A paper can do this by addressing the following questions:

\section{Introduction}
\label{sec:intro}

% Who is the \textbf{target audience}?

% What \textbf{problem} does the proposed system address?

% Why is the system important and \textbf{what is its impact}?

% How is the system \textbf{licensed}?

\begin{figure*}[ht!]
  \centering
  \includegraphics[width=1.0\linewidth]{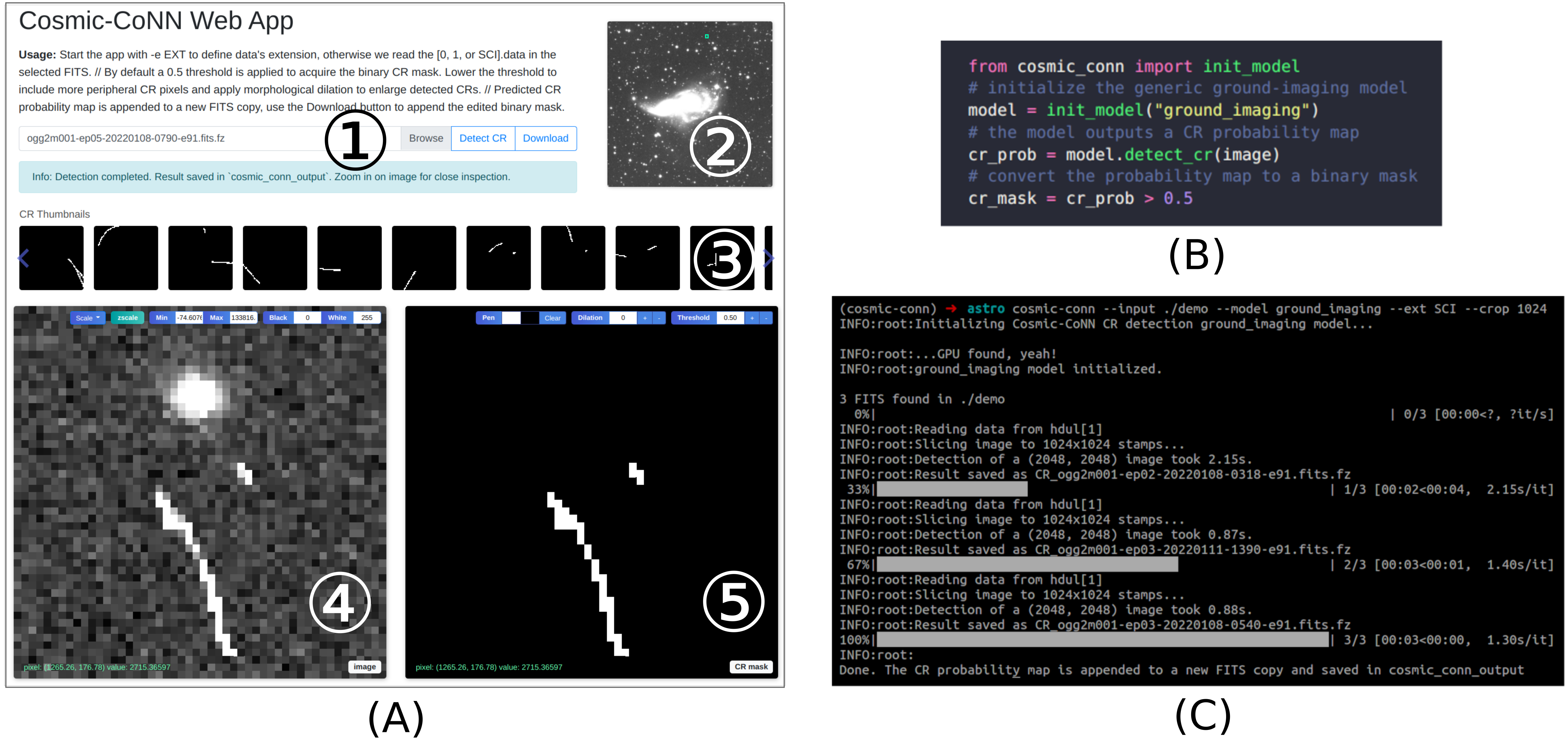}
   \caption{Our segmentation framework provides three user interfaces for different application scenarios: (A) the visualization toolkit has a graphical user interface (GUI) for model inference, interactive visualization, and mask editing, (B) clean Python library APIs for integration with user's data pipeline, and (C) a command-line interface for batch processing. GUI components: \CircledText{1} File input and output; \CircledText{2} Whole-image preview and navigation; \CircledText{3} Thumbnails shortcuts to detected objects ranked from large to small; \CircledText{4} Image window with various mapping (scale) algorithms and manual controls to visualize 16-bit floating point images; \CircledText{5} Segmentation mask window with synchronized field of view with the image. The highlighted pixels are detected CRs. The user can adjust the visualization of the HDR data on the left while interactively editing the segmentation mask on the right.}
   \label{fig:three_interfaces}
\end{figure*}

Semantic segmentation is not only a common computer vision task, but also a decades-old problem in astronomy. For astrophysicists whose research relies on observing the universe with optical telescopes and charge-coupled device (CCD) imagers, identifying cosmic rays (CRs) in their observations has been a challenging task\cite{Windhorst_1994, van_Dokkum_2001, groom2004nonsense, curtis_mccully_2018_1482019}. Telescope images 
can be a few megapixels or up to 3,200 megapixels \cite{10.1117/12.857920}, in contrast, CR-contaminated pixels are often just a few pixels wide. Because these bright pixels can be mistaken for real astronomical sources, it is necessary to reject them before further scientific interpretation of the data (see CR detection examples in Fig.~\ref{fig:three_interfaces} \CircledText{5}). 

% CRs are high-energy particles from outer space that can travel through the atmosphere and penetrate below ground. When these particles collide with telescopes' charge-coupled device (CCD) or active-pixel sensor (CMOS) imagers, they can excite electrons in image sensors and create artifacts of various shapes, shown in Fig.~\ref{fig:cr_shapes}. CR-contaminated pixels may vary in length and size depending on the thickness of the image sensor \cite{groom2004nonsense} but they are often just a few pixels wide, making them hard to detect, especially when only a single exposure is available.

% \begin{figure}[h]
%   \centering
%   \includegraphics[width=1.0\linewidth]{fig1_cr_shapes.pdf}
%   \caption{Three types of cosmic rays (CRs) found in the same image (reversed), marked in circles. Patch size in pixels. CRs may have various shapes and sizes but normally just a few pixels wide.}
%   \label{fig:cr_shapes}
% \end{figure}

Identifying tiny CRs in multi-megapixel images is only the first step. Astronomy telescope imagers are often cooled to operate below freezing temperature to minimize detector dark current and other noise sources \cite{Brown_2013}, and these highly sensitive CCD sensors produce 16-bit floating point high dynamic range (HDR) images that require special software for visualization. Without a scientific visualization tool that supports native integration with popular deep-learning frameworks, the detection and mask verification are divided into separate steps that involve exporting and reading files between different tools.

Given existing tools, the workflow of segmentation, image visualization, human inspection, and possible editing of the mask is a cumbersome process involving switching between multiple tools or software, making it worthwhile to develop a dedicated tool to streamline this workflow. In our \href{https://github.com/cy-xu/cosmic-conn}{video demonstration}, we show an interactive process that involves continuous adjustments to both the science image and the segmentation mask to acquire the accurate coverage of a CR that might affect the analysis of an adjacent stellar object. This level of seamless interaction was previously impossible if one were switching between different tools after each adjustment.

Computer vision researchers can integrate this visualization toolkit with other segmentation models to provide end users, especially domain experts who are not machine learning researchers, an interactive graphical user interface (GUI) (Fig.~\ref{fig:three_interfaces}) in production. The streamlined workflow enables the user to do real-time segmentation, HDR image visualization, and interactive mask inspection and editing without switching tools. The GUI toolkit allows any user to benefit from deep-learning-powered tools without having to know programming. The browser-based tool can be readily hosted on a graphics processing unit-ready (GPU) server so users in the private/public network can enjoy GPU acceleration from any device (Section~\ref{sec:multi-user}). 

In addition, future tiny-object or high-precision segmentation tasks can adopt the interactive interface as an annotation tool for pixel-level labeling in multi-megapixel images, especially for HDR data. The Python backend allows native integration with popular deep-learning frameworks, with the potential to be an interface for Active Machine Learning and Interactive Machine learning (Sec.~\ref{sec:discussion}). 

The discussion of detection algorithms is not the focus of this paper so we briefly introduce Cosmic-CoNN \cite{Xu2021}, our deep-learning segmentation framework deployed at Las Cumbres Observatory (LCO) for identifying cosmic rays in astronomical images. We curated a large, diverse dataset \cite{xu_chengyuan_2021_5034763} of over 4,500 scientific observations from LCO's 23 globally distributed telescopes\footnote{LCO has 25 telescopes around the world now. Our research, started in 2020, used data from all 23 then-operational instruments. \url{https://lco.global/}} \cite{Brown_2013}. In this dataset we discovered an extreme 1 to 10,000 class imbalance between CR and non-CR pixels that presented a challenge for previous machine-learning models. We proposed a novel loss function, and other improvements, to address this issue and increase model generalization. Our model achieved $99.91\%$ true-positive rate at a fixed false-positive rate of $0.01\%$ on LCO instruments and maintains over $96.40\%$ true-positive rate on data from another observatory, acquired with instruments that were never used for training (see \cite{Xu2021} for details). Our CR detector has become part of LCO's BANZAI data reduction pipeline that processes hundreds of astronomical observations every day \cite{curtis_mccully_2018_1257560}. 

\begin{figure*}[ht!]
  \centering
  \includegraphics[width=1.0\linewidth]{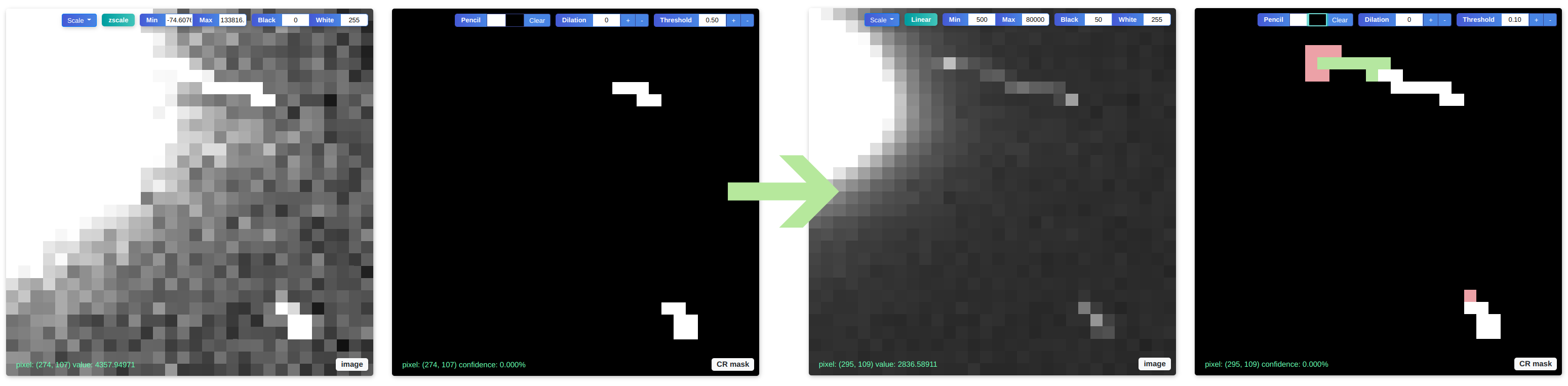}
   \caption{When the model or the default threshold does not produce ideal results (left), the user can adjust the HDR image mapping for better visualization, at the same time edit the mask interactively (right). The probability threshold and morphological dilation allow for global mask manipulation, while the pencil tool allows pixel-level mask editing. Pixels that are manually added/deleted by the user are marked in green or red, which will override the global manipulations.}
   \label{fig:tools}
\end{figure*}

Here we summarize the main contributions of our interactive visualization toolkit:
\begin{itemize}
    \item We provide a streamlined workflow for CR detection, improving quality by enabling human-in-the-loop segmentation, and reducing the overall time cost of astronomical image analysis and interpretation.
    \item We address a use case that is common in scientific imaging, but not well-supported by existing tools: interactive segmentation in large, multi-megapixel HDR images with tiny objects.
    \item We release our software as an open-source package that can be deployed off-the-shelf with diverse image types and segmentation models, and can facilitate imaging research across many scientific disciplines.
\end{itemize}

In addition, we found the tool useful during the development of our tiny-object segmentation model. The interactive visualization provides timely feedback for changes to the image processing pipeline, making it a useful research-support tool in computer vision as well.

% In addition, we as CV researchers found the tool helpful during the development of our tiny-object segmentation model. Being able to inspect the model's results closely and interactively, especially the tiny objects, without having to switch between the deep-learning pipeline and astronomical visualization tools encourages timely feedback to each of the changes we made to the image processing pipeline and even the neural network. This makes the visualization toolkit a useful research-support tool for computer-vision researchers as well.

\section{Usage}
\label{sec:usage}

% \textbf{How} does the system work?

The visualization toolkit shown in Fig.~\ref{fig:three_interfaces} A is the key component to unify model inference, image visualization, segmentation mask inspection and editing into a single interface. It can visualize 2-dimensional NumPy arrays \cite{harris2020array} and directly read FITS\footnote{FITS is an image and table format widely used for astronomical data \url{https://fits.gsfc.nasa.gov/fits_documentation.html}} files. It takes only 3 seconds to detect and render a 4-megapixel (2,000 by 2,000 pixels) 16-bit floating point image on a consumer laptop with a low-power NVIDIA RTX 3060 GPU.

The image window and segmentation mask window are always synchronized to an identical field of view (Fig.~\ref{fig:three_interfaces}~\CircledText{4}~\&~\CircledText{5}). This design provides a very useful reference for close inspection of tiny objects in large images. The user can navigate and zoom-in/out with mouse controls in any of the image windows, including the overview image~\CircledText{2}. Thumbnail shortcuts~\CircledText{3} allow the user to quickly jump to and inspect detected objects, making it a unique design especially useful for locating tiny objects in very large images.

The image window (Fig.~\ref{fig:tools}) provides multiple mapping algorithms to map (clip/normalize) 16-bit floating point data to 8-bit unsigned integers, including linear, logarithmic, and square-root scaling, as well as \href{https://docs.astropy.org/en/stable/api/astropy.visualization.ZScaleInterval.html#astropy.visualization.ZScaleInterval}{IRAF’s zscale}, an algorithm preferred by astronomers. The modular design of the image processing pipeline (Sec.~\ref{pipeline}) allows new mapping algorithms to be added easily. In addition, a user can manually assign the minimum and maximum range to read from raw data for the versatility especially needed in HDR images. The bottom left corner of each window shows the mouse cursor's pixel-location value in original data and the predicted mask's confidence.

In the segmentation mask window (Fig.~\ref{fig:tools}), a user can raise or lower the default $0.5$ threshold to acquire a binary mask from the deep-learning model's predicted probability map $\in [0, 1]$. The user can then apply morphological operations like dilation to manipulate the mask globally, or use the pencil tool to manually edit the mask at pixel-level. 

In the context of CR detection, the Download button will append the edited segmentation mask to the FITS file. This behavior can be changed based on the application. We can also change the communication mechanism with the deep-learning framework so the user can initiate the iterative labeling and training process in an active learning setting.

\begin{figure*}[ht!]
  \centering
  \includegraphics[width=0.9\linewidth]{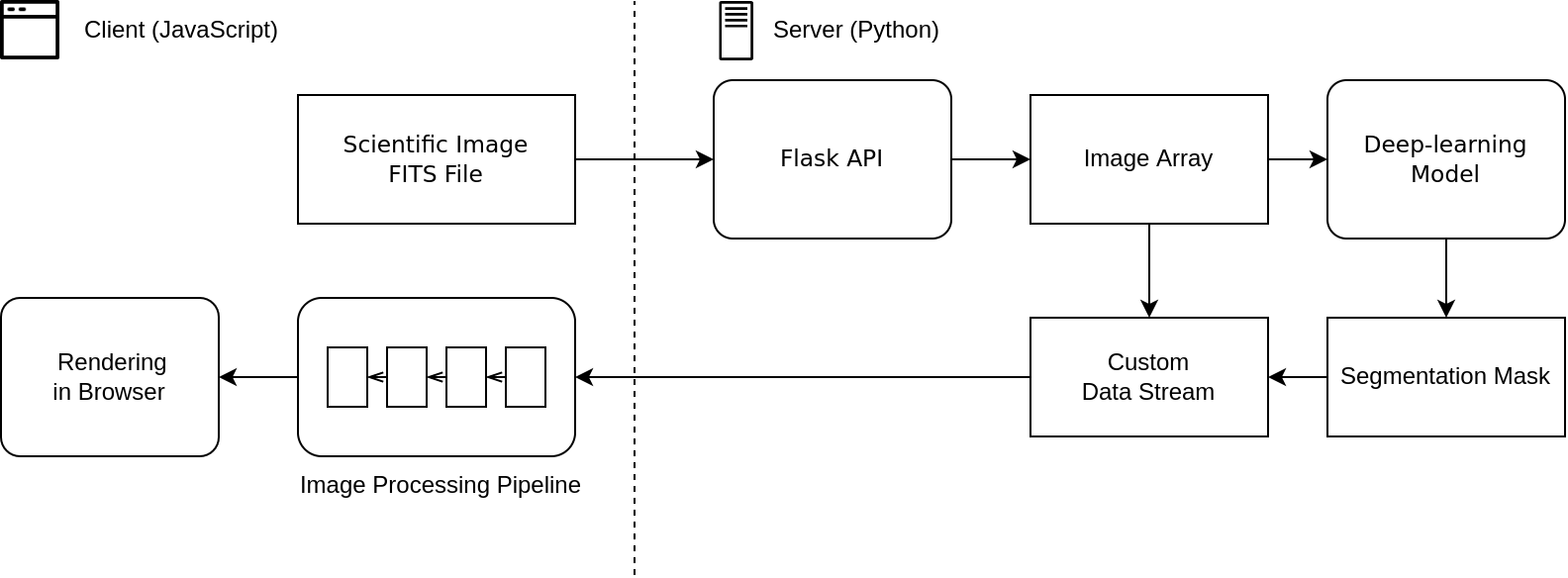}
   \caption{The interaction visualization toolkit's data flow architecture between the client and the server.}
   \label{fig:data_pipeline}
\end{figure*}

\section{System Design}\label{sec:system}

The visualization toolkit is powered by a Flask backend and JavaScript frontend (Fig.~\ref{fig:data_pipeline}). The Python-based backend allows seamless integration with popular deep-learning frameworks. We can run the server's instance locally or hosted on a cloud server for remote user access. The server-end only handles model inference and user instance management while the image processing pipeline happens entirely in the browser at the client-end. This design avoids overloading the server when hosted for multi-user access. The communication between the client and the server only happens at file uploading and downloading using a custom data steam to reduce the network delay.

\subsection{Image Processing Pipeline}\label{pipeline}

We adopt a modular design in the image processing pipeline to maximize the flexibility to add or remove image operations in the pipeline. The science image and the segmentation mask go through an ordered sequence of operations, and the modular design reduces computation and shortens the response time as the image is buffered after each operation -- an adjustment in the middle of the pipeline will only trigger later stages to reprocess the image.

In the context of astronomical data, the pipeline will first apply user's manual min-max clipping to the raw data, then apply a three-sigma clipping to remove outliers (over saturation and dead pixels). By default the previously mentioned zscale algorithm is applied to map the 16-bit image to 8-bit before rendering in the browser. 

The segmentation mask's pipeline is simpler as only one scalar threshold is applied to the probability mask to to acquire the binary mask. We use a separate mask to track the user's manual edits and combine with the binary mask before rendering in the browser.

\begin{figure*}[ht!]
  \centering
  \includegraphics[width=0.95\linewidth]{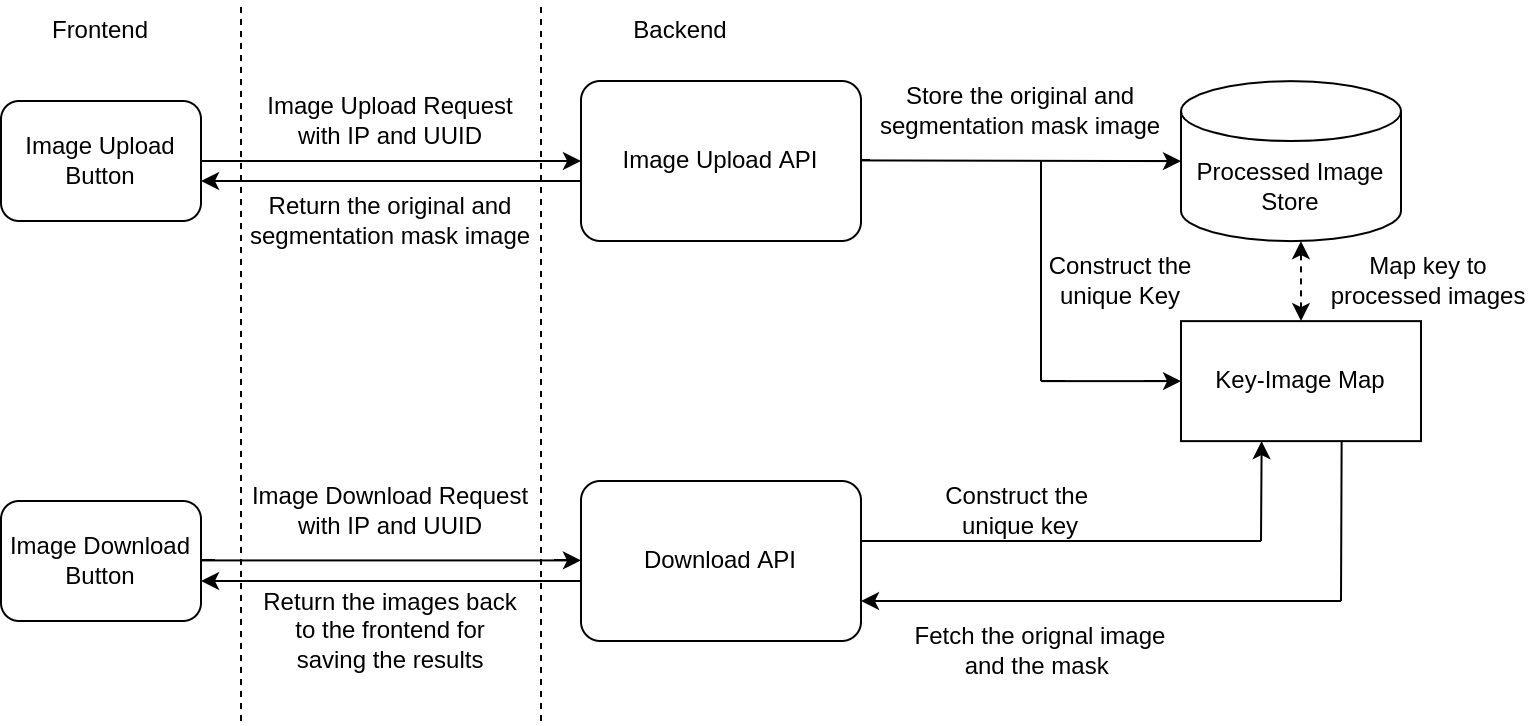}
   \caption{Frontend and backend architecture for multi-user interaction.}
   \label{fig:multiuser}
\end{figure*}

\subsection{Multi-user Support}\label{sec:multi-user}

Unlike many machine learning researchers, most of the real-world model end users do not have access to GPUs. With this in mind, we designed the GUI toolkit as a web-based application to support GPU acceleration and multiple-user access from any device. In a large-scale deployment, additional cloud GPU resources could be recruited as necessary to support a higher number of users.

Fig.~\ref{fig:multiuser} illustrates our system architecture, which supports multi-user interaction via secure user sessions. In addition to user's IP address, a universally unique identifier (UUID) is sent to the server to identify each detection request, either through the upload request or the download request. A unique request key is constructed and the server will maintain a map of the key and temporary path of the uploaded files with the segmentation mask appended. When the user is done with editing the image and requests to download the combined results, the key is used to retrieve the correct file. In this way, we avoid race conditions when multiple users interact with the same deployed application.

\section{Advantages Over Existing Tools}

% How does it \textbf{compare with existing systems}?
% What is the \textbf{novelty} in the approach/technology on which this system is based?

% We build the GUI toolkit's backend with Flask web application framework \cite{grinberg2018flask}. The lightweight yet powerful Python backend provides native support for PyTorch library and allows researchers to easily integrate with other popular deep-learning frameworks like Tensorflow \cite{tensorflow2015-whitepaper}. The browser-based tool can be hosted on the cloud or internal network with GPU acceleration to support multi-user accessing. The synchronized dual-window design and the thumbnail shortcuts are especially useful for inspecting and editing tiny objects in large masks.

Our interactive segmentation and visualization toolkit has the following key features:
\begin{itemize}
    \item A synchronized dual-window design (Fig.~\ref{fig:three_interfaces} A) and thumbnail-based image navigation (Fig.~\ref{fig:three_interfaces} \CircledText{3}) enable inspecting and editing tiny objects in large images;
    \item Computer vision researchers can inspect the results interactively via the GUI toolkit to better understand the model's behavior and assist their research. They can also deploy a GUI segmentation tool for end users in production environment with little effort;
    \item The browser-based application can be hosted on the cloud or internal GPU server to support multi-user access and GPU acceleration from any device;
    \item The Python and Flask backend allow seamless integration with popular deep-learning frameworks. Researchers can adapt this GUI for high-precision annotation or Active/Interactive Machine Learning.
\end{itemize}

SAOImageDS9 \cite{joye2005ds9} is a powerful FITS image visualization tool widely used in the astronomical community. DS9 inspired us to develop the GUI components in our toolkit. It supports various multi-frame layouts like tiling, blinking, and coordinates aligning. Despite active development, it remains primarily focused on visualization and we do not see an easy solution to integrate this standalone software with popular deep-learning frameworks.

ImageJ \cite{Schneider2012} is an image analysis program extensively used in the biological sciences. ImageJ2 \cite{Rueden2017} is a rewrite for multidimensional image data. It provides powerful image processing functionalities but requires a third-party plug-in to synchronize two image windows, and we haven't found a solution to make tiny object searching in large images as easy as the thumbnail shortcuts provided in our toolkit. ImageJ is versatile and general-purpose, but not optimized for the deep-learning segmentation workflow.

DeepImageJ \cite{deepimageJ_2021} is a plugin to support the use of pre-trained deep-learning models in ImageJ. It provides access to various models in a biomedical model repository (BioImage Model Zoo), and allows basic deep-learning model inference. But it also carries over ImageJ's disadvantages we discussed above and is hard to integrate with popular deep-learning frameworks, especially for researchers who need interactive data analysis during the research stage.

ITK-SNA \cite{py06nimg} is well known for 3D medical image segmentation, providing powerful functionalities from community contributions. But it lacks the support for deep-learning methods and the standalone software is hard to integrate with other frameworks. 

% We developed an interactive image segmentation and visualization framework for identifying, inspecting, and editing the mask of tiny objects to tackle the aforementioned CR detection problem in astronomical images. Our ene-to-end solution Cosmic-CoNN \cite{2021arXiv210614922X} features an interactive segmentation and visualization toolkit with out deep-learning framework 

\section{Discussion}\label{sec:discussion}

This demonstration highlights our three-in-one toolkit (segmentation, visualization, and editing) which streamlines the CR detection workflow and enables human-in-the-loop, interactive tiny-object segmentation in large, multi-megapixel, HDR images. In the future, we anticipate that user interfaces such as this one will be instrumental in the development of Interactive Machine Learning (IML) systems. Such systems are a promising approach for machine learning in domains where unlabeled data are abundant but annotations are expensive or difficult to obtain. The IML learning paradigm is especially beneficial in areas where domain knowledge is required, like biomedicine, astronomy, material science, etc., in which it is helpful for domain experts to steer the model training process. IML also reduces the overhead for scientists in various disciplines to train machine learning models \cite{Amershi2014Power}. Our interactive frontend and backend architecture is a step towards that direction.

Our dataset, CR detection model, and interactive visualization toolkit are open source and available at \url{https://github.com/cy-xu/cosmic-conn}. New features, such as instance segmentation and multi-file interface, are under consideration. We look forward to other computer vision researchers joining the open-source project to make this toolkit more useful for its various applications in astronomy, computer vision, interactive machine learning, and other research areas.

We appreciate the helpful discussion and feedback from Prof. Jennifer Jacobs, Jiaxiang Jiang, Alex Rich, Kuo-Chin Lien, and members from the Expressive Computation Lab of University of California, Santa Barbara.

%%%%%%%%% REFERENCES
{\small
\bibliographystyle{ieee_fullname}
\bibliography{main}

\begin{thebibliography}{10}\itemsep=-1pt

\bibitem{Amershi2014Power}
Saleema Amershi, Maya Cakmak, William~Bradley Knox, and Todd Kulesza.
\newblock Power to the people: The role of humans in interactive machine
  learning.
\newblock {\em AI Magazine}, 35(4):105--120, Dec. 2014.

\bibitem{Brown_2013}
T.~M. Brown, N. Baliber, F.~B. Bianco, M. Bowman, B. Burleson, P. Conway, M.
  Crellin, {\'{E}}. Depagne, J.~De Vera, B. Dilday, D. Dragomir, M. Dubberley,
  J.~D. Eastman, M. Elphick, M. Falarski, S. Foale, M. Ford, B.~J. Fulton, J.
  Garza, E.~L. Gomez, M. Graham, R. Greene, B. Haldeman, E. Hawkins, B.
  Haworth, R. Haynes, M. Hidas, A.~E. Hjelstrom, D.~A. Howell, J. Hygelund,
  T.~A. Lister, R. Lobdill, J. Martinez, D.~S. Mullins, M. Norbury, J. Parrent,
  R. Paulson, D.~L. Petry, A. Pickles, V. Posner, W.~E. Rosing, R. Ross, D.~J.
  Sand, E.~S. Saunders, J. Shobbrook, A. Shporer, R.~A. Street, D. Thomas, Y.
  Tsapras, J.~R. Tufts, S. Valenti, K.~Vander Horst, Z. Walker, G. White, and
  M. Willis.
\newblock Las cumbres observatory global telescope network.
\newblock {\em Publications of the Astronomical Society of the Pacific},
  125(931):1031--1055, sep 2013.

\bibitem{groom2004nonsense}
Don Groom.
\newblock Cosmic rays and other nonsense in astronomical ccd imagers.
\newblock In Paola Amico, James~W. Beletic, and Jenna~E. Beletic, editors, {\em
  Scientific Detectors for Astronomy}, pages 81--94, Dordrecht, 2004. Springer
  Netherlands.

\bibitem{deepimageJ_2021}
Estibaliz Gómez-de Mariscal, Carlos García-López-de Haro, Wei Ouyang,
  Laurène Donati, Emma Lundberg, Michael Unser, Arrate Muñoz-Barrutia, and
  Daniel Sage.
\newblock Deepimagej: A user-friendly environment to run deep learning models
  in imagej.
\newblock {\em Nature Methods}, 18(10):1192–1195, 2021.

\bibitem{harris2020array}
Charles~R. Harris, K.~Jarrod Millman, St{\'{e}}fan~J. van~der Walt, Ralf
  Gommers, Pauli Virtanen, David Cournapeau, Eric Wieser, Julian Taylor,
  Sebastian Berg, Nathaniel~J. Smith, Robert Kern, Matti Picus, Stephan Hoyer,
  Marten~H. van Kerkwijk, Matthew Brett, Allan Haldane, Jaime~Fern{\'{a}}ndez
  del R{\'{i}}o, Mark Wiebe, Pearu Peterson, Pierre G{\'{e}}rard-Marchant,
  Kevin Sheppard, Tyler Reddy, Warren Weckesser, Hameer Abbasi, Christoph
  Gohlke, and Travis~E. Oliphant.
\newblock Array programming with {NumPy}.
\newblock {\em Nature}, 585(7825):357--362, Sept. 2020.

\bibitem{joye2005ds9}
W.~A. {Joye} and E. {Mandel}.
\newblock {The Development of SAOImage DS9: Lessons Learned from a Small but
  Successful Software Project}.
\newblock In P. {Shopbell}, M. {Britton}, and R. {Ebert}, editors, {\em
  Astronomical Data Analysis Software and Systems XIV}, volume 347 of {\em
  Astronomical Society of the Pacific Conference Series}, page 110, Dec. 2005.

\bibitem{10.1117/12.857920}
S.~M. Kahn, N. Kurita, K. Gilmore, M. Nordby, P. O'Connor, R. Schindler, J.
  Oliver, R.~Van Berg, S. Olivier, V. Riot, P. Antilogus, T. Schalk, M. Huffer,
  G. Bowden, J. Singal, and M. Foss.
\newblock {Design and development of the 3.2 gigapixel camera for the Large
  Synoptic Survey Telescope}.
\newblock In Ian~S. McLean, Suzanne~K. Ramsay, and Hideki Takami, editors, {\em
  Ground-based and Airborne Instrumentation for Astronomy III}, volume 7735,
  pages 257 -- 273. International Society for Optics and Photonics, SPIE, 2010.

\bibitem{curtis_mccully_2018_1482019}
Curtis McCully, Steve Crawford, Gabor Kovacs, Erik Tollerud, Edward Betts,
  Larry Bradley, Matt Craig, James Turner, Ole Streicher, Brigitta Sipocz,
  Thomas Robitaille, and Christoph Deil.
\newblock astropy/astroscrappy: v1.0.5 zenodo release, Nov. 2018.

\bibitem{curtis_mccully_2018_1257560}
Curtis McCully, Monica Turner, N Volgenau, Daniel Harbeck, Stefano Valenti,
  Austin Riba, Etienne Bachelet, Ira~W. Snyder, Brodie Kurczynski, Martin
  Norbury, and Rachel Street.
\newblock Lcogt/banzai: Initial release, June 2018.

\bibitem{Rueden2017}
Curtis~T. Rueden, Johannes Schindelin, Mark~C. Hiner, Barry~E. DeZonia,
  Alison~E. Walter, Ellen~T. Arena, and Kevin~W. Eliceiri.
\newblock Imagej2: Imagej for the next generation of scientific image data.
\newblock {\em BMC Bioinformatics}, 18:1--26, 11 2017.

\bibitem{Schneider2012}
Caroline~A Schneider, Wayne~S Rasband, and Kevin~W Eliceiri.
\newblock Nih image to imagej: 25 years of image analysis, 2012.

\bibitem{van_Dokkum_2001}
Pieter~G. van Dokkum.
\newblock Cosmic-ray rejection by laplacian edge detection.
\newblock {\em Publications of the Astronomical Society of the Pacific},
  113(789):1420--1427, nov 2001.

\bibitem{Windhorst_1994}
Rogier~A. Windhorst, Barbara~E. Franklin, and Lyman~W. Neuschaefer.
\newblock Removing cosmic-ray hits from multiorbit {HST} wide field camera
  images.
\newblock {\em Publications of the Astronomical Society of the Pacific},
  106:798, jul 1994.

\bibitem{Xu2021}
Chengyuan Xu, Curtis Mccully, Boning Dong, D~Andrew Howell, and Pradeep Sen.
\newblock Cosmic-conn: A cosmic ray detection deep-learning framework, dataset,
  and toolkit, 2021.

\bibitem{xu_chengyuan_2021_5034763}
Chengyuan Xu, Curtis McCully, Boning Dong, D.~Andrew Howell, and Pradeep Sen.
\newblock Cosmic-conn lco cr dataset, June 2021.

\bibitem{py06nimg}
Paul~A. Yushkevich, Joseph Piven, Heather Cody~Hazlett, Rachel Gimpel~Smith,
  Sean Ho, James~C. Gee, and Guido Gerig.
\newblock User-guided {3D} active contour segmentation of anatomical
  structures: Significantly improved efficiency and reliability.
\newblock {\em Neuroimage}, 31(3):1116--1128, 2006.

\end{thebibliography}
}

\end{document}